%% file: neurips_2024.tex
\documentclass{article}

\usepackage[preprint]{neurips_2024}

\usepackage[utf8]{inputenc} 
\usepackage[T1]{fontenc}    
\usepackage{hyperref}       
\usepackage{url}            
\usepackage{booktabs}       
\usepackage{amsfonts}       
\usepackage{nicefrac}       
\usepackage{microtype}      
\usepackage{xcolor}         
\usepackage{graphicx}
\usepackage{amsmath}
\usepackage{multirow}
\usepackage{makecell}
\usepackage{cleveref}
\usepackage{wrapfig}
\usepackage{microtype}
\usepackage{authblk}

\newcommand{\mypar}[1]{\vspace{-0.8em}\paragraph{#1}}
\newcommand{\gangli}[1]{}
\newcommand{\lyc}[1]{}

\def\eg{e.g.}
\def\ie{i.e.}
\def\upvspacefig{}

\makeatletter
  \newcommand\figcaption{\def\@captype{figure}\caption}
  \newcommand\tabcaption{\def\@captype{table}\caption}
\makeatother

\title{SpeechForensics: Audio-Visual Speech Representation Learning for Face Forgery Detection}

\author{
  \textbf{Yachao Liang}\textsuperscript{\textmd{1},\textmd{2}}\quad
  \textbf{Min Yu}\textsuperscript{\textmd{1},\textmd{2}}\thanks{Corresponding authors}\quad
  \textbf{Gang Li}\textsuperscript{\textmd{3}}\quad
  \textbf{Jianguo Jiang}\textsuperscript{\textmd{1},\textmd{2}}\quad
  \textbf{Boquan Li}\textsuperscript{\textmd{4}$*$}\quad\\
  \vspace{-0.8em}
  \textbf{Feng Yu}\textsuperscript{\textmd{5}}\quad
  \textbf{Ning Zhang}\textsuperscript{\textmd{6}}\quad
  \textbf{Xiang Meng}\textsuperscript{\textmd{1},\textmd{2}}\quad
  \textbf{Weiqing Huang}\textsuperscript{\textmd{1},\textmd{2}}\\
  \textsuperscript{\textmd{1}}Institute of Information Engineering, Chinese Academy of Sciences\\
  \textsuperscript{\textmd{2}}School of Cyber Security, University of Chinese Academy of Sciences\\
  \textsuperscript{\textmd{3}}Deakin University\quad
  \textsuperscript{\textmd{4}}Harbin Engineering University\\
  \textsuperscript{\textmd{5}}Institute of Computing Technology, Chinese Academy of Sciences\\
  \textsuperscript{\textmd{6}}Institute of Forensic Science, Ministry of Public Security\\
  \texttt{\{liangyachao, yumin\}@iie.ac.cn}
}

\begin{document}

\maketitle

\input{tex/abstract}

\input{tex/intro}
\input{tex/relatedwork}
\input{tex/methods}
\input{tex/experiments}

\input{tex/conclusions}
\input{tex/acknowledge}

{
    \small
    \bibliographystyle{plain}
    \bibliography{mybib}
}

\input{tex/X_suppl}


\end{document}

%% file: tex/abstract.tex
\begin{abstract}
Detection of face forgery videos remains 
a formidable challenge in the field of digital forensics,
especially the generalization to unseen datasets and common perturbations.
In this paper,
we tackle this issue 
by leveraging the synergy between audio and visual speech elements, embarking on a novel approach through 
audio-visual speech representation learning.
Our work is motivated by the finding that audio signals,
enriched with speech content, 
can provide precise information effectively reflecting facial movements.
To this end,
we first learn precise audio-visual speech representations on real videos via a self-supervised masked prediction task,
which encodes both local and global semantic information simultaneously.
Then, 
the derived model is directly transferred to the forgery detection task.
Extensive experiments demonstrate that 
our method outperforms the state-of-the-art methods 
in terms of cross-dataset generalization and robustness,
without the participation of any fake video in model training.
The code is available \href{https://github.com/Eleven4AI/SpeechForensics}{\textcolor{red}{here}}.
\end{abstract}

%% file: tex/intro.tex
\section{Introduction}\label{sec-intro}

The rapid advancement of generative models enables 
synthetic realistic facial images~\cite{creswell2018generative,kingma2013auto,karras2019style},
and they have significantly enhanced face manipulation techniques, 
allowing for the replacement of facial identities and 
the modification of attributes such as 
expressions~\cite{thies2016face2face,thies2019deferred}  
and lip movements\cite{deepfakesurl,faceswapurl}.
While these advancements offer vast potential 
for entertainment and filmmaking, 
they also harbor the risk of misuse for deceptive purposes.

In response to these concerns, 
there has been a surge in the development of 
face forgery detection methodologies grounded in deep learning
\cite{afchar2018mesonet,rossler2019faceforensics++,
wang2022deepfake,zheng2021exploring,li2020face,
zhao2021learning,haliassos2021lips,wang2023altfreezing}.
Despite these efforts, 
it is widely acknowledged that 
face forgery detectors frequently experience a decline in effectiveness 
when confronted with novel manipulation techniques~\cite{cozzolino2018forensictransfer,chai2020makes,li2020celeb}. 
This vulnerability poses significant hurdles to 
the reliable application of these detection systems, 
highlighting a critical area for ongoing research and innovation.

To enhance the generalization capabilities of face forgery detectors,
researchers have proposed various methods aimed at 
mining more discriminative clues~\cite{zhao2021learning,wang2021representative,zheng2021exploring,haliassos2022leveraging,wang2023altfreezing}.
Some works focus on detecting spatial artifacts left 
in the process of facial manipulation~\cite{chai2020makes,li2020face,zhao2021learning,bai2023aunet},
especially blending boundaries~\cite{li2020face,shiohara2022detecting}.
However,
these methods are sensitive to common perturbations,
making them difficult to generalize to real-life scenarios.
Another line of research resorts to model temporal features~\cite{zeng2021contrastive,haliassos2022leveraging,zhao2022self,wang2023altfreezing},
considering that fake videos are synthesized in a frame-by-frame manner.
They identify unnatural facial movements existing in fake videos by applying special architectures~\cite{zheng2021exploring}
or introducing auxiliary tasks~\cite{ciftci2020fakecatcher,haliassos2021lips,haliassos2022leveraging}.
Although showing promising results,
short-term information
modeling capacity
(\eg,
1 second~\cite{zheng2021exploring,haliassos2021lips})
makes them overfit to specific low-level temporal features to varying degrees,
resulting in their suboptimal generalization on unseen datasets and perturbations,
as observed in our experiments.

This motivates us to find more general semantic-level features to detect anomalous facial movements.
Recent efforts on audio-visual speech recognition have shown that 
accurate speech contents can be extracted from both audio signals 
and lip movements~\cite{shi2022avhubert,haliassos2023jointly,ma2023auto}.
Inspired by this,
we conjecture that audio signals could provide strong semantic supervision
for identifying inaccurate lip movements in fake videos,
given that lip sequences and audio segments in a real video 
should convey the same speech contents.
This brings us to the key problem:
\emph{how to extract semantically rich speech-related features to represent detailed lip movements?}

An intuitive solution is to align the speech representations of each frame of audio segments and lip sequences directly,
as in~\cite{chung2017out}.
However,
this method will fail to detect fake videos processed 
by lip synchronization techniques,
such as \texttt{Wav2Lip}~\cite{prajwal2020lip},
commonly used by recent talking face generation technologies.
Considering that local lip synchronization cannot bring long-range temporal coherence,
we further propose to perform forgery detection 
by learning audio and visual speech representations in a framework
that encodes both phonetic and linguistic information,
which we term as local and global information.
Specifically,
it learns local information by frame-wisely audio-visual representation alignment
and models global dependencies via masked prediction task,
following previous speech representation learning methods~\cite{shi2022avhubert,haliassos2023jointly,zhu2023vatlm}.
In this way,
both short-range and long-range temporal features are learned.
After learning audio-visual speech representation on real videos,
we directly transfer the trained model to the forgery detection task 
by finding discrepancies 
between visual and audio speech representations in fake videos.

Thanks to the unsupervised manner and high-level semantic learning,
our method,
termed \texttt{SpeechForensics},
avoids overfitting on forgery features and shows strong robustness on various perturbations.
We conduct comprehensive experiments 
to evaluate the effectiveness of our method,
and it shows strong performance under different manipulations, datasets,
and perturbations.
Especially,
our method achieves the AUC of 99.0\% 
on FakeAVCeleb~\cite{khalid2021fakeavceleb} and 91.7\% on KoDF~\cite{kwon2021kodf}.
Our main contributions are summarized as follows:
\begin{itemize}

\item We propose to perform face forgery video detection by extracting speech representations from audio and visual streams.
It learns on real videos and can smoothly transfer to the forgery detection task,
markedly streamlining the forgery detection workflow.

\item 
We demonstrate a simple framework,
which encodes both short-range and long-range temporal information,
is well-suited to our method.
And we tailor it to a face forgery detector using the proposed modality alignment module.

\item Extensive experiments demonstrate the superiority of our method over the state-of-the-art methods in terms of cross-dataset generalization,
robustness,
and interpretability,
in an unsupervised manner.
\end{itemize}

%% file: tex/relatedwork.tex
\section{Related Work}\label{sec-relatedwork}

\mypar{Face Forgery Detection.}
Initial approaches in face forgery detection predominantly treat the task 
as a binary classification problem, 
leveraging deep learning models trained on datasets 
specifically compiled for detecting forgeries~\cite{afchar2018mesonet, rossler2019faceforensics++, dang2020detection, li2021frequency}.
For instance,
\cite{afchar2018mesonet} introduces a pair of detection networks 
known as \texttt{Mesonet} and \texttt{MesoInception},
demonstrating that 
lightweight neural networks can effectively 
undertake forgery detection tasks.
Analogously,
\cite{rossler2019faceforensics++} highlights 
the superior performance of 
an unconstrained \texttt{Xception} network over its predecessors, 
focusing primarily on the analysis of spatial details 
within individual frames.
Subsequently,
some works~\cite{guera2018deepfake, amerini2019deepfake,zhang2021detecting, gu2022region, de2020deepfake, gu2022delving}
try to combine temporal networks to perform forgery detection.
Despite the promising results in the in-dataset setting,
these vanilla methods usually suffer from severe performance degradation 
when facing unseen forgeries.

\mypar{General Forgery Detection.}
To boost the generalization of detectors on unseen forgeries,
researchers attempt to find more discriminative features 
at both image and video levels.

Image-based methods~\cite{chai2020makes,li2020face,zhao2021learning,shiohara2022detecting,bai2023aunet}
analyze the spatial artifacts common to forged faces and generate synthetic data to guide models to focus on them.
For example,
\texttt{Face X-ray}~\cite{li2020face} and \texttt{SBI}~\cite{shiohara2022detecting} detect blending boundaries
 caused by the fusion of the forged face and background,
 and \texttt{AUNet}~\cite{bai2023aunet} concentrates on the relation between different facial action units.
 While they are adept at identifying specific artifacts,
 these artifacts are easily destroyed by some common perturbations,
 \eg, compression,
 which makes it difficult to generalize to real scenarios.

On another front, 
video-based methods make efforts to explore temporal clues
 via special network architectures~\cite{zheng2021exploring,wang2023altfreezing}
  or auxiliary tasks~\cite{li2018ictu,haliassos2021lips,haliassos2022leveraging,zhao2022self}.
\texttt{FTCN}~\cite{zheng2021exploring} reduces the convolutional kernel size to 1
 forcing the network to only focuses on temporal features.
\texttt{LipForensics}~\cite{haliassos2021lips} uncovers unnatural lip movements
 via pre-training on the lipreading task and finetuning on forgery datasets.
 Analogously,
 \texttt{RealForensics}~\cite{haliassos2022leveraging} leverages cross-modal self-supervision learning to capture facial movements.
 Despite their notable performance,
 they tend to rely on short-range low-level temporal features,
 leading to their limited generalization on new datasets and robustness against common perturbations.
 In contrast,
 our method detects both short-range and long-range anomalous facial movements using semantic-level information and functions in an unsupervised manner,
 inherently possessing superior generalization and robustness.

\mypar{Audio-Visual Speech Representation Learning.}
The advent of extensive large-scale audio-visual speech datasets 
\cite{afouras2018deep, afouras2018lrs3, chung2018voxceleb2}
has spurred the development of numerous audio-visual speech representation learning methods 
in recent years~\cite{chung2017out, prajwal2020lip, shi2022avhubert, ma2023auto}.
\cite{chung2017out} learns visual and audio speech representations simultaneously 
based on synchronization signals between lip movements and corresponding audio segments.
Benefiting from off-the-shelf audio speech recognition models,
\cite{ma2021lira} proposes to learn visual speech representations 
by minimizing the distance 
between learned visual embeddings and pre-trained audio embeddings.
\cite{haliassos2023jointly} further advances this field 
by simultaneously learning visual and auditory speech representations 
through student-teacher networks.  
We choose \texttt{AVHuBERT}~\cite{shi2022avhubert} 
as the implementation of our audio-visual speech representation learning,
considering it fits into our framework,
\ie, learning both local and global semantic representations,
and demonstrates remarkable efficacy in downstream speech recognition tasks.

%% file: tex/methods.tex
\section{Method}\label{sec-methods}
Our method consists of the audio-visual speech representation learning stage and the face forgery detection stage.
We first learn semantically rich visual and audio speech representations in a unified feature space from real videos,
which can be implemented by many audio-visual speech representation learning approaches~\cite{shi2022avhubert,zhu2023vatlm,haliassos2023jointly}.
Subsequently, 
the model leverages these representations to pinpoint discrepancies 
between lip movements and corresponding audio segments in fake videos. 

\subsection{Speech Representation Learning}

In order to simultaneously learn local and global speech information,
strongly correlated with short-range and long-range lip movements,
we conduct frame-wise audio-visual representation alignment and the masked prediction task.
Since we mainly focus on the forgery detection task,
we will briefly introduce the representation learning stage,
for details refer to~\cite{shi2022avhubert}.

\begin{figure*}[t]
  \vspace{-1.5em}
  \centering
  \includegraphics[width=0.99\linewidth]{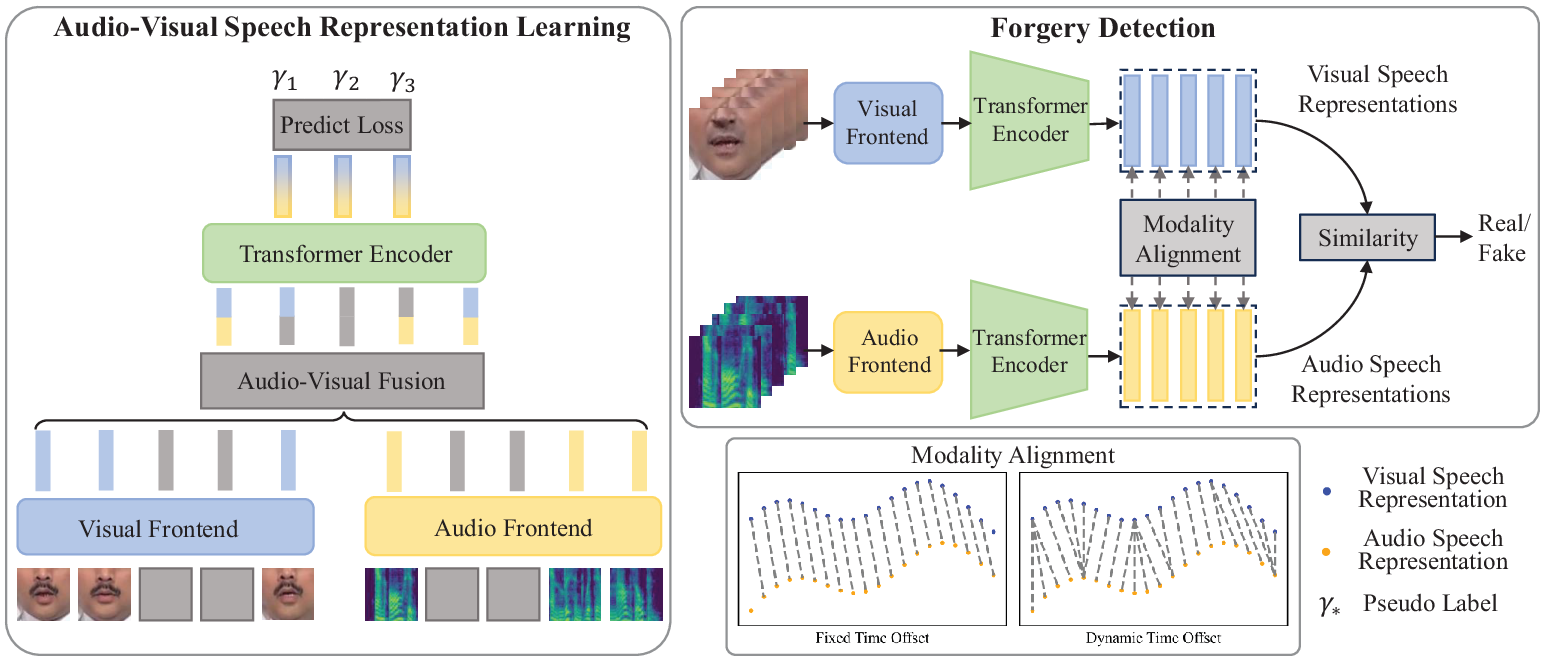}
  \begin{flushleft}
        \vspace{-3mm}
         \vspace{-3mm}
  \end{flushleft}  
    \caption{\textbf{Overview of the proposed method.}
    During the stage of audio-visual speech representation learning,
    local speech representations and global information are learned by frame-wise feature alignment and the masked prediction task,
    respectively.
    In the stage of forgery detection,
    we separately feed the whole lip movement sequence and audio stream of a video into the learned model to get visual and audio speech representations.
    And we flag videos with low matching scores between visual and audio speech representations as fake videos.
    }
    \label{fig:method}
    \vspace{-1.5em}
\end{figure*}

\mypar{Local representation alignment.}
Considering
$\mathcal{X}=\{(\boldsymbol{I^{i}},\boldsymbol{A^{i}})\}^{N}_{i=1}$ 
as the set of audio-visual pairs extracted from real videos.
Given a visual and audio pair 
$(\boldsymbol{I}_{1:T},\boldsymbol{A}_{1:T})$ from the set $\mathcal{X}$,
where $T$ represents the sequence length of clip.
\gangli{you didn't mention clip at all.}
We first get intermediate features 
$F^{v}_{1:T}=f^{v}_{e}(\boldsymbol{I}_{1:T})$ and 
$F^{a}_{1:T}=f^{a}_{e}(\boldsymbol{A}_{1:T})$ 
through the visual frontend $f^{v}_{e}$ and the audio frontend $f^{a}_{e}$, 
respectively.
And $F^{v}$ and $F^{a}$ are fused by channel-wise concatenation,
before which modality dropout is applied to allow the unimodal input.
Then the fused features are fed into subsequent transformer encoder to learn frame-wise speech representations,
as shown in \cref{fig:method}.
The target labels for training are derived 
through cluster assignment~\cite{hsu2021hubert}, 
which are initialized based on~\texttt{MFCC} features of audio and iteratively refined with audio-visual features learned by encoders via k-means.
And we denote them as $\boldsymbol{\gamma}_{1:T}\in\{1,2,\dots,C\}$,
where $C$ is the size of the codebook.
By this means,
the representations of every frames of visual and audio modalities are aligned in a unified feature space.

\mypar{Global information modeling.}
Following the described procedure, 
the visual and audio speech representations for each frame are synchronized, 
facilitating the computation of their similarity in subsequent analyses.
However,
local speech contents conveyed by lip movements,
i.e.,
visemes,
only contain limited temporal features and can easily be tampered by lip-sync methods,
\eg,
\texttt{Wav2Lip}~\cite{prajwal2020lip}.
To address this problem,
we further introduce global temporal information modeling.

We employ the masked-prediction task,
a method extensively utilized across various fields~\cite{kenton2019bert,kegler2020deep,cai2023marlin},
to model contextual dependencies effectively.
Let $M_{v},M_{a}\subset\{1,2,\dots,T\}$ 
represent the sets of indices of visual and audio masked sequences, 
this task can be formulated as:
\begin{equation}
    \mathcal{L}=\sum_{t \in M_{v} \cup M_{a}}\log p(\gamma_{t} \mid \tilde{F}^{v},\tilde{F}^{a})
\end{equation}
where $\tilde{F}^{a}$ and $\tilde{F}^{v}$ 
denotes of the corrupted audio features and visual features,
respectively.

\subsection{Face Forgery Detection}
For the forgery detection,
we aim to  identify discrepancies 
between visual and audio speech representations in manipulated videos.
\gangli{or forgery videos? or are you sure it is manipulated? or just in videos?}  
To achieve this,
we extract visual embeddings as visual speech representations 
by feeding the learned model with only visual inputs,
and apply the same procedure for audio embeddings.
Notably, 
obtaining accurate speech representations from the audio stream is more straightforward, 
allowing these to function effectively as pseudo-labels.

Given the visual and audio speech representations of any video,
we get their matching score by calculating frame-wise cosine similarity,
which can be formulated as:
\begin{equation}
    \mathcal{S}(e^{v},e^{a})=\frac{1}{T}\sum_{t=1}^{T}sim(e^{v}_{t},e^{a}_{t})
    \label{equ:similarity}
\end{equation}
where $e_{v}$ and $e_{a}$ represent the visual and audio embeddings 
extracted from the final layer of our model,
respectively,
and $sim(\cdot,\cdot)$ is the cosine similarity between two vectors.
And videos exhibiting low matching scores are consequently classified as forgeries.

Another crucial problem is the time offsets between visual and audio signals,
which inevitably exist even in real videos 
due to recording or encoding errors~\cite{afouras2022self,feng2023self}.
To address this problem,
we consider two time offset assumptions,
\ie,
fixed and dynamic,
and introduce different modality alignment approaches to alleviate their impact.
See the intuitive illustration of these two assumptions in \cref{fig:method}.

\mypar{Assumption 1: Fixed time offset.}
First,
we assume that the frame offsets between visual and audio streams remain constant throughout the video.
We apply a sliding-window technique,
as in~\cite{chung2017out},
to correct time offsets prior to calculating final matching scores.
Specifically,
based on the assumption that 
the maximum offset between visual and audio streams is $\tau$,
we compute the cosine similarity 
between the feature of each visual frame and 
its adjacent audio features within a window of $\pm\tau$ frames.
Thereafter,
the highest average cosine similarity across this window
is taken as the overall similarity of the video.
Consequently, 
the similarity~\cref{equ:similarity} can be re-formulated as:
\begin{equation}
    \mathcal{S}(e^{v},e^{a})=\max_{t-\tau \leq i \leq t+\tau} \frac{1}{T} \sum_{t=1}^{T}sim(e^{v}_{t},e^{a}_{i})
\end{equation}
where we pad zero vectors as $e^{a}_{i}$ when $i<1$ or $i>T$.

\mypar{Assumption 2: Dynamic time offset.}
On the contrary,
we also consider the dynamic assumption,
\ie,
the frame offsets between visual and audio signals vary over time.
Based on this assumption,
we introduce the \emph{Dynamic Time Warping} (DTW) algorithm~\cite{muller2007dynamic}, 
commonly used to align and calculate the similarity of two time series data.
In this case,
\cref{equ:similarity} will be re-written as:
\begin{equation}
    \mathcal{S}(e^{v},e^{a})=DTW(e^{v}_{1:T},e^{a}_{1:T})
\end{equation} 
where we also apply cosine similarity as the cost measure of DTW.   

%% file: tex/experiments.tex
\section{Experiments}\label{sec-experiments}
\subsection{Experimental Setup}
\vspace{0.2em}
\mypar{Dataset.}
We evaluate our methods across three distinct video forgery datasets:
\texttt{Faceforensics++} (FF++)~\cite{rossler2019faceforensics++},
\texttt{FakeAVCeleb}~\cite{khalid2021fakeavceleb} and \texttt{KoDF}~\cite{kwon2021kodf}.
Note that Only the FakeAVCeleb contains videos belong to the Real-Visual-Fake-Audio category,
and we exclude them as we focus on the facial forgery and to maintain fairness of experiments.

\textbf{Faceforensics++} contains 1,000 real videos alongside 4,000 fake videos,
created via four different manipulation methods. 
These include face swapping methods 
(\texttt{DeepFakes}~\cite{deepfakesurl}, \texttt{FaceSwap}~\cite{faceswapurl}), 
and two face reenactment methods 
(\texttt{Face2Face}~\cite{thies2016face2face} and \texttt{NeuralTextures}~\cite{thies2019deferred}).
For our evaluation, 
We re-download videos using the provided YouTube IDs and extract audio segments 
from the provided frame locations. 
Contrary to the common practice of treating \texttt{Faceforensics++} 
as a visual-only dataset, 
we paired the original videos with their corresponding audio segments 
to create an audio-visual test dataset. 
After excluding videos unavailable or containing non-corresponding mouth movements and voices,
we selected 500 videos from each category for testing.

\textbf{FakeAVCeleb} contains 500 real videos and 19,500 fake videos.
It is derived from \texttt{VoxCeleb2}~\cite{chung2018voxceleb2} 
and represents diverse ethnic backgrounds,
ages and genders.
This dataset involves four manipulation techniques,
\texttt{Faceswap}~\cite{korshunova2017fast} and 
\texttt{Faceswap GAN} (\texttt{FSGAN}) ~\cite{nirkin2019fsgan} for face swapping,
\texttt{SV2TTS}~\cite{jia2018transfer} for real-time cloning voice, 
and \texttt{Wav2Lip}~\cite{prajwal2020lip} for audio-driven facial reenactment.

\textbf{KoDF}~\cite{kwon2021kodf} is a large-scale Korean forgery datasets,
containing 62,166 real videos and 175,776 fake videos.
For our evaluation, 
we focus on fake videos crafted using four distinct manipulation techniques: 
\texttt{FaceSwap}~\cite{faceswapurl}, 
\texttt{DeepFaceLab}~\cite{perov2020deepfacelab}, 
\texttt{FOMM}~\cite{siarohin2019first}, 
\texttt{Audio-driven} (including ATFHP~\cite{yi2020audio} 
and \texttt{Wav2Lip}~\cite{prajwal2020lip}). 
From each of these categories, 
we randomly select 1,000 videos to compile our testing set.

\mypar{Preprocessing.}
We first utilize FFmpeg~\cite{tomar2006converting} to convert all videos into 25fps and audio into 16kHz sample rate.
For each video clip, 
we initiate the process by identifying faces using \texttt{RetinaFace}~\cite{deng2020retinaface} 
and subsequently extract facial landmarks with \texttt{FAN}~\cite{bulat2017far}. 
We then align the frames using affine transformations, 
and crop $96\times96$ regions centered around the mouth,
as indicated by the landmarks.
For the audio part,
we extract \texttt{MFCC} features from wavform every 10ms,
and we concatenate 4 adjacent audio frames to align with visual modality.

\mypar{Architecture \& Training.}
The audio-visual speech representation model consists of visual frontend,
audio frontend and masked predictor.
Following AVHuBert~\cite{shi2022avhubert},
We use the Resnet-18 2D+3D~\cite{ma2021end} as the visual frontend.
And the audio frontend contains only a single linear projection layer to avoid the over-reliance issue on the audio stream~\cite{shi2022avhubert}.
The masked predictor is implemented by the standard transformer encoder.
Further details can be found in~\cref{appendix:architecture}.

The model is trained on 
\texttt{LRS3}~\cite{afouras2018lrs3} and \texttt{VoxCeleb2}~\cite{chung2018voxceleb2} datasets,
which contain 433 and 1326 hours of videos respectively.
The training process adhere to the methodology outlined by~\citep{shi2022avhubert}.
For the purp-
oses of our experiments, 
unless otherwise noted, 
we utilize a publicly available pretrained model~\footnote{\url{https://github.com/facebookresearch/av_hubert}}. 

\subsection{Quantitative Comparisons}
We compare our method with several state-of-the-art detectors,
including \texttt{Xception}~\cite{rossler2019faceforensics++},
\texttt{Patch-based}~\cite{chai2020makes},
\texttt{Face X-ray}~\cite{li2020face},
\texttt{LipForensics}~\cite{haliassos2021lips},
\texttt{FTCN}~\cite{zheng2021exploring},
\texttt{RealForensics}~\cite{haliassos2022leveraging} and
\texttt{AVAD}~\cite{chen2022self}.
Moreover,
we also construct a model,
consisting of two Resnet 2D~\cite{he2016deep} models,
to extract only phonetic-level (5 frames) speech information for comparison,
And we dub it \texttt{SpeechForensics-Local}.
See~\cref{appendix:baselines} for more details about above detectors.

We undertake extensive experiments focusing on the generalization and robustness of detectors.
In line with established practices in the field~\cite{haliassos2021lips, haliassos2022leveraging, shiohara2022detecting, bai2023aunet, wang2023altfreezing},
we utilize the area under the receiver operating characteristic curve (AUC) 
to gauge the efficacy of our method at the video level. 
Unless otherwise specified,
we take the fixed time offset assumption.

\mypar{Cross-Manipulation Generalization.}
\label{cross-manipulation}
In real-world situations,
detectors frequently encounter novel manipulation techniques, 
underscoring the necessity for these systems to 
possess robust generalization capabilities against unseen manipulations.
To assess our method's ability to generalize across different manipulation methods, 
we conducted evaluations on the widely recognized FF++ high-quality (HQ) dataset.
For supervised methods,
the experiments adopt the leave-one-out strategy,
in line with established practices~\cite{haliassos2021lips, haliassos2022leveraging, zheng2021exploring}.
It is noteworthy that the settings for both cross-manipulation and cross-dataset  are equivalent for unsupervised methods,
\ie,
AVAD~\cite{feng2023self} and our method.

\input{table/cross_manipulation}
The AUC results presented in \cref{table:cross_manipulation}
demonstrate that 
our method  either matches or exceeds performance across different categories, 
notably without utilizing any forgery samples. 
Remarkably, our approach achieves perfect results (100\%) 
with the two face reenactment methods, 
\texttt{Face2Face} and \texttt{NeuralTextures}. 
However, 
the performance on \texttt{FaceSwap} (91.1\%) is slightly lower 
compared to the other categories, 
a trend that aligns with findings from related methods 
such as \texttt{LipForensics} and \texttt{RealForensics}~\cite{haliassos2022leveraging}).
This could be attributed to \texttt{FaceSwap}'s use of target face landmarks 
for generating source faces, 
which might result in more precise lip shapes. 
Nonetheless, 
our method demonstrates the significant capability 
in detecting such forgeries through contextual analysis, 
indicating its effectiveness against diverse manipulation techniques.
Notably, 
our method significantly outperforms \texttt{AVAD},
which nearly produces random results.
And we note that \texttt{SpeechForensics-Local} also achieves considerable performance on this dataset,
although modeling local speech information.

\mypar{Cross-Dataset Generalization.}
We  further extend our evaluation to include a cross-dataset comparison, 
aligning with the practise in prior works~\cite{haliassos2021lips, haliassos2022leveraging, chen2022self}.
This involves testing the performance of our method 
on the unseen datasets \texttt{FakeAVCeleb}~\cite{khalid2021fakeavceleb} and \texttt{KoDF}\cite{kwon2021kodf},
with the supervised models initially trained on the \texttt{FF}++ dataset.
In addition,
we also report the results of every category within the \texttt{FakeAVCeleb} dataset,
which is segmented into five categories based on the manipulation techniques used
\texttt{Faceswap}~\cite{korshunova2017fast} (FS), 
\texttt{FSGAN}~\cite{nirkin2019fsgan}, 
\texttt{Wav2Lip}~\cite{prajwal2020lip} (WL), 
\texttt{Faceswap-Wav2Lip} (FS-WL) and \texttt{FSGAN-Wav2Lip} (FSGAN-WL),
with the latter two categories indicating the combined use of manipulation methods. 

\input{table/cross_dataset}
The results in \cref{table:cross_dataset} show that 
our method significantly outperforms both supervised and unsupervised counterparts 
in the cross-dataset setting,
outperforming previous state-of-the-art method,
RealForensics,
by 8.8\% on FakeAVCeleb and 7.4\% on KoDF.
It is worth noting that 
\texttt{SpeechForensics-Local} fails to detect fake videos generated by \texttt{Wav2Lip},
as we mentioned above,
suggesting the key role of global temporal information for forgery video detection.
Conversely,
\texttt{AVAD}~\cite{feng2023self} shows promise in 
detecting forgeries generated by \texttt{Wav2Lip}~\cite{prajwal2020lip},
though its performance on other types of forgery  
is yet to be fully assessed.
Our approach, 
in contrast, 
delivers exceptional performance across all forgery types, 
achieving a perfect 100\% on \texttt{Wav2Lip}-manipulated video. 
This underscores the fact that accurate lip synchronization at a local level 
does not necessarily imply global semantic integrity.
We also provide more multimodal experiment results in~\cref{appendix:more_comparisons}.

\mypar{Cross-Language Generalization.}
Considering the linguistic diversity encountered in real-world video content, 
we expand our evaluation to 
assess the cross-language generalization capabilities of our approach. 
To categorize the languages present in the FF++ dataset,
we  utilize \texttt{Whisper}~\cite{radford2023robust} for language detection 
and subsequently split the videos into various language categories:
including \emph{English} (EN), 
\emph{Arabic} (AR), \emph{Spanish} (ES), \emph{Russian} (RU), 
\emph{Ukrainian} (UK), \emph{Tagalog} (TL) and others.
The results, presented in~\cref{table:cross_language}, 
illustrate the AUC scores achieved for each language category. 
Our findings indicate that our method maintains effective performance across different languages, 
\input{table/cross_language}
even though it was originally trained on datasets predominantly in \emph{English}. 
This outcome underscores the versatility of the audio-visual speech representations 
learned by our model, 
demonstrating their language-agnostic nature and 
highlighting the method's potential applicability in diverse linguistic contexts.

\mypar{Robustness to Unseen Perturbations.}
Considering the prevalence of image post-processing operations on social media platforms, 
such as compression, 
the robustness of detection systems emerges as a crucial challenge.
In line with previous studies~\cite{haliassos2021lips, haliassos2022leveraging, wang2023altfreezing, feng2023self},
we evaluate the robustness of our method against various perturbations in the FF++ dataset,
and these include Color saturation change, 
Color contrast change, Block-wise distortion, Gaussian noise, 
Gaussian blur, Pixelation and Video compression,
each applied at 5 different intensity levels as in~\cite{jiang2020deeperforensics}.

\begin{figure*}[t]
  \centering
  \upvspacefig
  \includegraphics[width=0.9\linewidth]{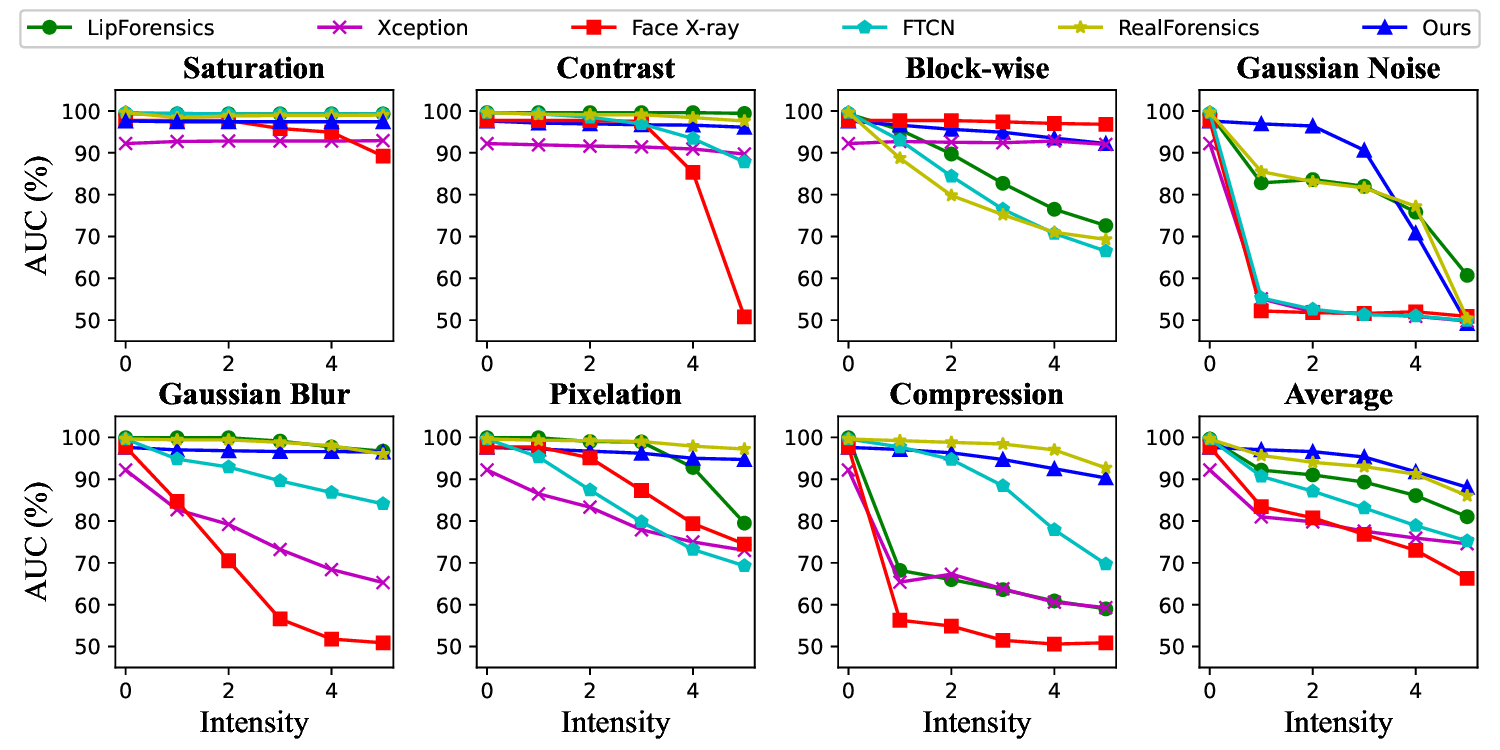}
  \vspace{-0.5em}
  \caption{\textbf{Robustness to unseen perturbations.} Video-level AUC scores (\%) are reported under different perturbations.
  Each perturbation contains five intensity levels~\cite{jiang2020deeperforensics}.
  ``Average'' denotes the mean of each perturbation under each intensity level.}
  \label{fig:robustness}
  \vspace{-0.5em}
\end{figure*}

The results in \cref{fig:robustness} 
compare the performance of our method against five supervised baselines under these conditions.
It can be found that 
detectors primarily relying on low-level texture features,
such as  \texttt{Face X-ray}~\cite{li2020face}, 
\texttt{Xception}~\cite{rossler2019faceforensics++} and \texttt{FTCN}~\cite{zheng2021exploring},
are vulnerable to most of the common perturbations. 
Notably, 
while \texttt{Face X-ray} and \texttt{Xception} show diminished effectiveness 
against perturbations that attenuate high-frequency content 
(e.g., \emph{blur}, \emph{compression}), 
\texttt{FTCN} is particularly sensitive to 
perturbations that disrupt temporal coherence (e.g., \emph{noise}).

Besides,
the three video-based methods,
FTCN~\cite{zheng2021exploring},
LipForensics~\cite{haliassos2021lips} and RealForensics~\cite{haliassos2022leveraging},
have all shown sensitivity to Block-wise distortion,
suggesting they rely on low-level temporal features to some extent.
Conversely,
our method demonstrates exceptional robustness against all types of perturbations.
Moreover,
our approach consistently outperforms \texttt{LipForensics},
which also models lip movements,
across various corruption scenarios,
although starting from a lower point (97.6\% vs 99.6\%).
This indicates that 
our method is capable of harnessing more potent semantic representations 
for the purpose of forgery detection, 
despite being trained exclusively on real data.
We provide perturbation examples and more results in~\cref{appendix:robustness}.

\subsection{Qualitative Results}
\begin{figure}[t]
    \centering
    \upvspacefig
    \includegraphics[width=\linewidth]{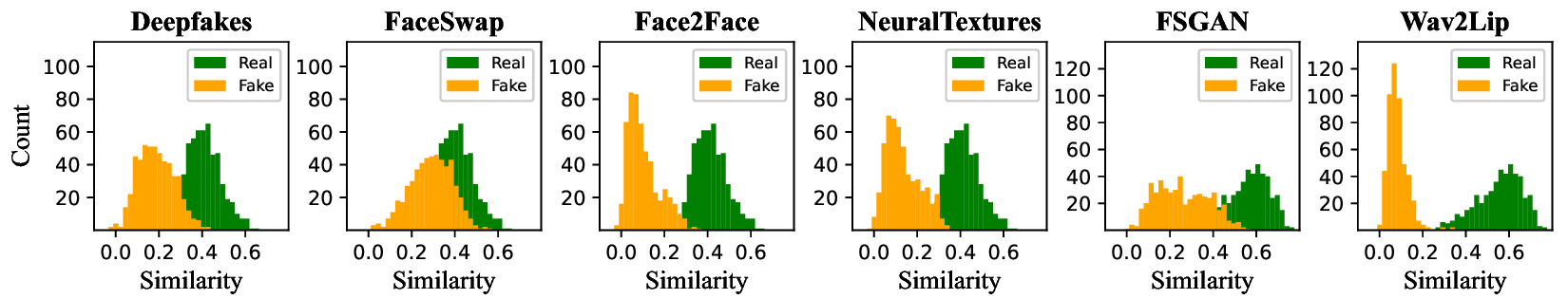}
    \vspace{-1.5em}
    \caption{\textbf{Visualized analysis.}
    Cosine similarity distributions of audio and visual speech representations for real videos and fake videos generated by different manipulation methods.}
    \label{fig:visual_distribution}
\end{figure}

\vspace{0.2em}
\mypar{Visualization.}
To demonstrate the effectiveness of our method,
we conduct an in-depth visual analysis utilizing \texttt{FF}++ and \texttt{FakeAVCeleb} datasets.
Specifically, 
we included all forgery categories from FF++,
\ie, \texttt{Deepfakes}, \texttt{FaceSwap}, \texttt{Face2Face} and \texttt{NeuralTextures}.
From the \texttt{FakeAVCeleb} dataset, 
we focused on manipulations made with \texttt{FSGAN} and \texttt{Wav2Lip}, 
selecting a random sample of 500 videos.  
As a result,
our experiment covers a total of six types of forgery.

For each type of forgery,
\gangli{sometimes you call it type, sometimes category. be consistent. }
\lyc{I use type for forgeries and category for datasets.}
\gangli{not the optimal way}
we calculate the cosine similarity 
between visual and corresponding audio speech representations 
extracted from videos.
\Cref{fig:visual_distribution} shows 
the cosine similarity distribution for each category.
As we can see,
almost all types of fake videos are clearly differentiated from real videos 
around a cosine similarity threshold of 0.3.  
An interesting finding is that 
our method exhibits exceptional performance on face reenactment techniques,
\ie, \texttt{Face2Face}, \texttt{NeuralTextures} and \texttt{Wav2Lip},
likely due to these methods prioritizing overall visual fidelity 
at the expense of accurate lip movements. 
More visualized results are in~\cref{appendix:visualization}.

\begin{figure*}[tb]
    \centering
    \upvspacefig
    \includegraphics[width=\linewidth]{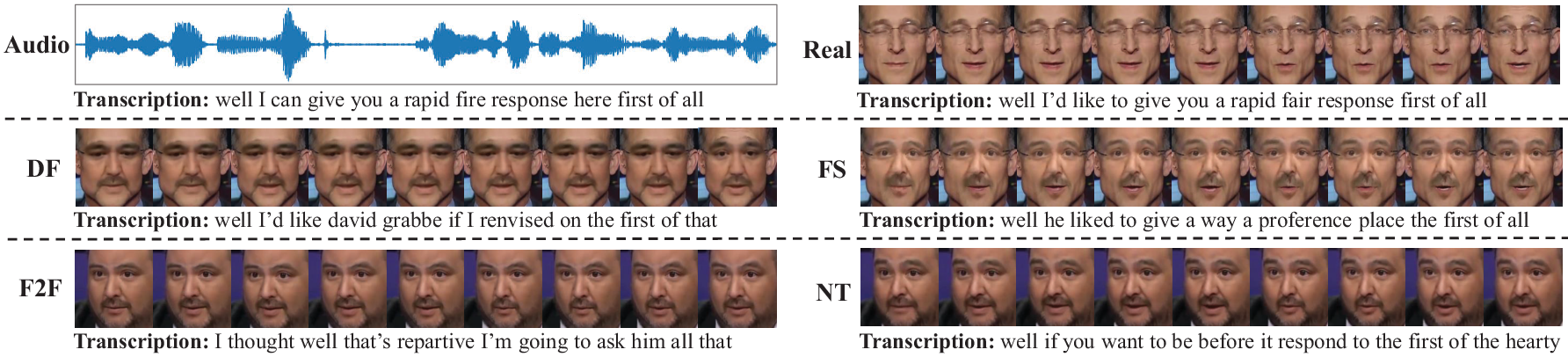}
    \vspace{-1.5em}
    \caption{\textbf{Interpretative analysis.}
    The transcriptions are based on audio and visual speech representations of real and different types of fake videos.
    We show the transcriptions of each type of video containing the same audio.}
    \label{fig:interpretability}
\end{figure*}
\mypar{Interpretative Analysis.}
Exploring another intriguing aspect, 
we delve into understanding how our method functions and 
what the derived audio-visual representations signify. 
To shed light on this, we conduct an interpretative analysis on the \texttt{FF}++.
Specifically,
we transcribe the extracted representations 
using an audio-visual speech recognition model as proposed by~\cite{shi2022avhubert},
fine-tuned for lipreading with the pretrained audio-visual speech representation model.
This enabled us to transcribe specific speech content from 
both the lip movements and corresponding audio segments in real and fake videos,
as shown in \cref{fig:interpretability}.
Note only the mouth region sequences are fed into the model to get the transcriptions.
\gangli{I rephrase above, please check whether they are appropriate. }
As can be seen,
the sentence transcribed from real lip movement frames  
is close to the sentence transcribed from audio. 
In contrast, 
lip movements in fake videos frequently result in nonsensical or chaotic transcriptions.
It suggests that 
these audio and visual speech representations indeed 
contain semantic information, 
which can be used for forgery detection.

\subsection{Ablation study}

\begin{figure}
\begin{minipage}{0.5\linewidth}
    \vspace{-1.5em}
    \centering
    \upvspacefig
    \includegraphics[width=0.98\linewidth]{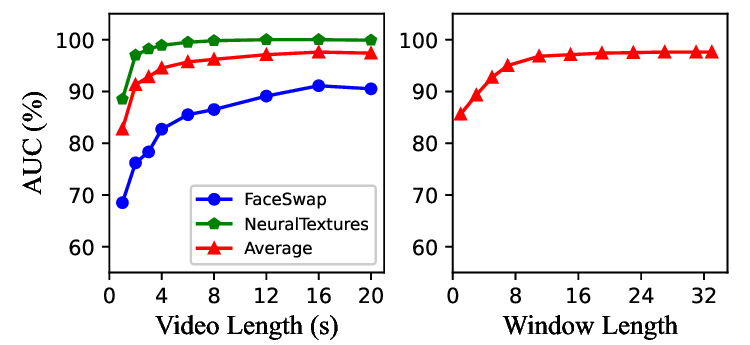}
    \figcaption{\textbf{Influence of video length and sliding-window length.}
    We evaluate the performance of our method conditioned on different input lengths and sliding-window lengths. }
    \label{fig:video_length}
    \vspace{-1.5em}
\end{minipage}
\begin{minipage}{0.5\linewidth}
    \vspace{-1.5em}
    \tabcaption{\textbf{Effect of different models and time offset assumptions.}
    We report the performance of models with different architectures and training datasets on FF++ and FakeAVCeleb.}
    \label{table:ablation_model}
    \vspace{1.5em}
    \centering
    \resizebox{0.98\linewidth}{!}{
    \begin{tabular}{l l l l c c }
    \toprule
    Model & Offset & Backbone & Dataset & FF++ & FakeAVCeleb \\  
    \midrule
    \multirow{5}{*}{AVHuBERT~\cite{shi2022avhubert}} 
        & \multirow{4}{*}{Fixed} 
            & BASE & LRS3 & 95.3 & 97.0 \\
            & & BASE & LRS3+Vox2 & 96.1 & 97.9 \\
            & & LARGE & LRS3 & 95.7 & 96.8 \\
            & & LARGE & LRS3+Vox2 & \textbf{97.6} & 99.0 \\
        \cmidrule(lr){2-6}
     & Dynamic & LARGE & LRS3+Vox2 & 93.7 & 98.7 \\
    \midrule
    VATLM~\cite{zhu2023vatlm} & Fixed & LARGE & LRS3+Vox2 & 97.1 & \textbf{99.3} \\
    \bottomrule
    \end{tabular}
    }
    \vspace{-1.5em}
\end{minipage}
\end{figure}

\vspace{0.2em}
\mypar{Influence of Video Clip Length.}
Given that our model accommodates video clips of varying lengths, 
we investigated how this aspect impacts the performance of our method on the FF++ dataset.
For this purpose, 
we selected \texttt{FaceSwap} and \texttt{Face2Face} as two emblematic types of forgeries 
and segmented videos into various durations, 
ranging from 1 second to 20 seconds, 
while maintaining consistency in all other hyperparameters. 
Results in \cref{fig:video_length} 
indicate a clear trend: 
the performance of our method is continuously enhanced with the extension of the video length to 16 seconds,
suggesting long-range temporal inconsistencies exist in forgery videos.
While previous detectors,
\eg,FTCN~\cite{zheng2021exploring} and RealForensics~\cite{haliassos2022leveraging},
can only utilize short-range temporal features,
resulting in their suboptimal performance.

\mypar{Different time offset assumptions.}
We also study the effect of different time offset assumptions between audio and visual streams,
\ie, fixed and dynamic.
As shown in \cref{table:ablation_model},
the dynamic based method achieves slightly better performance based on the fixed time offset assumption.
A reasonable explanation is that,
compared to real videos,
the time offsets of fake videos will be more inconsistent due to the uncertainty of the forgery process.
And the DTW algorithm makes fake videos have higher matching scores,
which is unfavorable to the forgery detection.
Furthermore,
we investigate the influence of different maximum offset $\tau$,
corresponding sliding-window length $2\tau+1$.
As~\cref{fig:video_length} shows,
increasing window length brings improved performance,
reaching a maximum of around 31.

\mypar{Different Models and Datasets.}
We further evaluate the effect of different models and training datasets.
For AVHuBERT~\cite{shi2022avhubert},
we use models with two configurations: 
BASE, which comprises 12 transformer blocks,
and LARGE, which includes 24 transformer blocks. 
Each configuration was trained on two datasets: 
\texttt{LRS3} alone and a combination of \texttt{LRS3} and \texttt{VoxCeleb2}, 
respectively.
Furthermore,
we evaluate another model,
VATLM~\cite{zhu2023vatlm},
which also fits our framework but incorporates text modality.
The experimental results on \texttt{FF}++ and \texttt{FakeAVCeleb} 
are shown in \cref{table:ablation_model}.
Results show larger models and more training data both boost the performance of our approach.
While compared with model size,
the influence introduced by datasets is more pronounced.
And both AVHuBERT and VATLM obtain remarkable results.

%% file: table/cross_manipulation.tex
\begin{wraptable}{r}{7.2cm}
\vspace{-1.5em}
\caption{\textbf{Cross-manipulation generalization.}
We report video-level AUC (\%) on FF++,
which contains four manipulation methods,
\ie, Deepfakes (DF), FaceSwap (FS), Face2Face (F2F) and NeuralTextures (NT).
$^*$ denotes results of our reproduction.
}
\label{table:cross_manipulation}
\centering
\resizebox{\linewidth}{!}{
\begin{tabular}{c l c c c c c}
\toprule
\multicolumn{2}{c}{\multirow{2}{*}{Method}} & \multicolumn{4}{c}{Train on remaining three} \\  
\cmidrule(lr){3-6}
\multicolumn{2}{c}{} & DF & FS & F2F & NT & \textbf{Avg}  \\
\midrule
\multirow{6}{*}{Supervised} 
        & Xception \cite{rossler2019faceforensics++} & 93.9 & 51.2 & 86.8 & 79.7 & 77.9  \\
        & Patch-based \cite{chai2020makes} & 94.0 & 60.5 & 87.3 & 84.8 & 81.7 \\
        & Face X-ray \cite{li2020face} & 99.5 & 93.2 & 94.5 & 92.5 & 94.9 \\ 
        & LipForensics \cite{haliassos2021lips} & 99.7 & 90.1 & 99.7 & 99.1 & 97.1 \\
        & RealForensics$^*$ \cite{haliassos2022leveraging} & \textbf{100.} & 96.1 & 99.5 & 97.0 & 98.1 \\
        & FTCN~\cite{zheng2021exploring} & 99.8 & \textbf{99.6} & 98.2 & 95.6 & \textbf{98.3} \\
\midrule
\multirow{3}{*}{Unsupervised} 
        & AVAD$^*$ \cite{feng2023self} & 59.2 & 55.1 & 59.9 & 58.4 & 58.2 \\
        & SpeechForensics-Local & 95.6 & 74.9 & 95.1 & 89.1 & 88.7 \\
        & SpeechForensics (ours) & 99.4 & 91.1 & \textbf{100.} & \textbf{100.} & 97.6 \\
\bottomrule
\end{tabular}
}
\vspace{-1.5em}
\end{wraptable}

%% file: table/cross_dataset.tex
\begin{table*}[h]
\vspace{-1.em}
\caption{\textbf{Cross-dataset generalization.} Video-level AUC (\%) on FakeAVCeleb and KoDF.
We report the results of every categories of FakeAVCeleb,
and the overall performance on it is reported in \textbf{Overall}.
The average performance over two datasets is reported in \textbf{Avg}.
}
\label{table:cross_dataset}
\centering
\resizebox{0.9\linewidth}{!}{
\begin{tabular}{llcccccccc}
\toprule
\multicolumn{2}{c}{\multirow{2}{*}{Method}} & \multicolumn{6}{c}{FakeAVCeleb} & \multirow{2}{*}{KoDF} & \multirow{2}{*}{\textbf{Avg}}\\  
\cmidrule(lr){3-8}
\multicolumn{2}{c}{} & FS & FSGAN & WL & FS-WL & FSGAN-WL & \textbf{Overall} &  & \\
\midrule
\multirow{6}{*}{Supervised} 
        & Xception \cite{rossler2019faceforensics++} & 67.0 & 62.5 & 59.7 & 57.2 & 68.0 & 61.6 & 77.7 & 69.7\\
        & Patch-based \cite{chai2020makes} & 97.4 & 80.5 & 78.9 & 93.8 & 87.8 & 83.6 & 83.9 & 83.8 \\
        & Face X-ray \cite{li2020face} & 89.9 & 85.4 & 69.5 & 84.4 & 87.6 & 78.4  & 83.0 & 80.7 \\ 
        & LipForensics \cite{haliassos2021lips} & 89.5 & 96.4 & 85.6 & 87.2 & 95.8 & 89.8 & 59.6 & 74.7 \\
        & FTCN\cite{zheng2021exploring} & 89.3 & 79.9 & 80.6 & 85.2 & 86.1 & 82.3 & 76.5 & 79.4 \\
        & RealForensics \cite{haliassos2022leveraging} & \textbf{98.1} & \textbf{100.} & 81.0 & 94.7 & 99.2 & 90.2 & 84.3  & 87.3 \\
\midrule
\multirow{3}{*}{Unsupervised} 
        & AVAD \cite{feng2023self} & 52.8 & 53.9 & 93.9 & 95.8 & 94.3 & 85.0 & 58.0 & 71.5  \\
        & SpeechForensics-Local & 69.3 & 85.4 & 0.10 & 0.08 & 0.08 & 19.0 & 48.3 & 33.7 \\
        & SpeechForensics (ours) & 93.9 & 96.0 & \textbf{100.} & \textbf{99.9} & \textbf{99.9} & \textbf{99.0} & \textbf{91.7} & \textbf{95.4} \\
\bottomrule
\end{tabular}
}
\end{table*}

%% file: table/cross_language.tex
\begin{wraptable}{r}{8cm}
\vspace{-0.5em}
\caption{\textbf{Cross-language generalization.}
AUC (\%) scores on videos of different languages in the FF++.}
\label{table:cross_language}
\centering
\resizebox{8cm}{!}{
\begin{tabular}{c c c c c c c c }
\toprule
Language & EN & AR & ES & RU & UK & TL & Others \\  
\midrule
AUC & 97.8 & 98.3 & 98.3 & 100. & 99.3 & 99.7 & 97.2 \\
\bottomrule
\end{tabular}
}
\vspace{-0.5em}
\end{wraptable}

%% file: tex/conclusions.tex
\section{Conclusion and Discussion} \label{sec-conclusions}
In this paper,
we have developed a method for forgery detection that 
identifies discrepancies between audio and visual speech representations. 
Demonstrating exceptional generalization capabilities to 
unseen manipulations and robustness against prevalent perturbations, 
our approach sets a new benchmark, 
notably without relying on fake videos for training.  
It also eliminates the need for finetuning and downstream tasks, 
significantly streamlining the detection workflow. 
Moreover,
since our method is based on speech representation learning,
it can be implemented in a training-free manner and may achieve better performance as the latter advances.
We are optimistic that 
our contributions will inspire further advancements in the field of forgery detection research.

\mypar{Limitations.}
While our method exhibits robust performance across diverse evaluations, 
it is not without its limitations. 
Primarily, 
it is constrained by its reliance on visual speech representations derived from lip movements, 
rendering it unsuitable for detecting forgeries that do not alter mouth regions. 
However, we note that facial forgeries typically involve the mouth area.
In addition,
it may suffer a certain level of performance degradation when encountering extreme testing samples,
\eg, videos with numerous silent clips or audio signals containing significant amount of ambient noise,
e.g.,
background music.
These considerations highlight areas for potential improvement and future exploration 
in enhancing the versatility and applicability of forgery detection techniques.

\mypar{Broader Impacts.}
Our work is aimed at fighting against face forgery technologies.
And we hope it could encourage more future detection works.
However,
since forgery and detection are two game-playing technologies, the emergence of new detection methods may lead to the evolution of forgery methods.
And we suggest that detection systems integrate different detection methods to combat potential new face forgery methods.

%% file: tex/acknowledge.tex
\section{Acknowledgement}
This work has been supported by the National Key R\&D Program of China (Grant No. 2021YFF0602104).

%% file: tex/X_suppl.tex
\newpage
\appendix
\section{Appendix}

\input{table/suppl_visual_frontend}
\subsection{Architecture Details.}
\label{appendix:architecture}
The visual frontend is a modified ResNet-18 \cite{ma2021end},
where the first convolutional layer is substituted by a 3D convolutional layer.
The resulting visual features are then transformed into 1024-dimensional tensors 
for each input frame through the application of 2D global average pooling. 
See \cref{table:suppl_visual_frontend} for more details.
The masked predictor consists of 24 standard transformer encoder blocks,
each featuring 16 attention heads and 1024 channels.
A final linear projection layer, with an output dimension of 256, is employed to deduce the ultimate cluster assignments.

\subsection{Compared Baselines}
\label{appendix:baselines}
We compare our method with both supervised and unsupervised methods.
\mypar{Supervised Methods.}
The state-of-the-art supervised methods 
include: 
1) \textbf{Xception}~\cite{rossler2019faceforensics++}: a widely used baseline for generalization comparison.
2) \textbf{Patch-based}~\cite{chai2020makes}: a patch-based forgery detection model with local receptive fields.
3) \textbf{Face X-ray}~\cite{li2020face}: a detector focusing on blending boundaries in fake images.
4) \textbf{LipForensics}~\cite{haliassos2021lips}: it targets unnatural mouth movements existing in forgery videos.
5) \textbf{FTCN}~\cite{zheng2021exploring}: a video detector modeling temporal features via special architecture.
6) \textbf{RealForensics}~\cite{haliassos2022leveraging}: 
it aims to learn temporally dense representations of facial movements via audio-visual self-supervision,
which facilitates the generalization of forgery detectors.

\mypar{Unsupervised Methods.}
1)\textbf{AVAD}~\cite{chen2022self}:
it leverages a pre-trained audio-visual synchronization network 
as its core framework, 
and adopts a downstream autoregressive model to 
learn the distribution of time delays between visual and auditory signals. 
2)\textbf{SpeechForensics-Local}: We utilize two Resnet 2D~\cite{he2016deep} models to extract audio and visual speech representations,
respectively.
The visual input contains 5 successive frames
and the audio encoder extract features from 0.2s audio clips.
We also train it on the~\texttt{LRS3}~\cite{afouras2018lrs3} dataset,
and we dub it as~\texttt{SpeechForensics-Local},
since it learns local phoneme information.

\begin{figure*}[h]
    \centering
    \upvspacefig
    \includegraphics[width=0.9\linewidth]{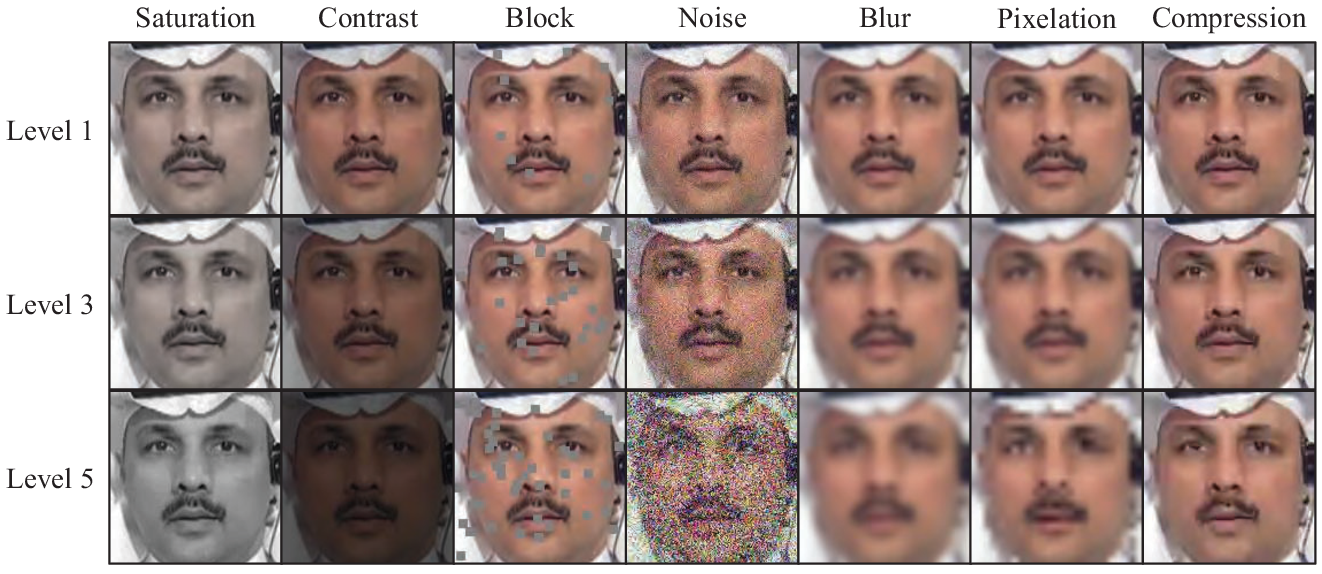}
    \caption{\textbf{Perturbed examples.}
    Visualization of all types of perturbations at different intensity levels.
    We present three representative (mild, moderate and severe) intensity levels.
    }
    \label{fig:suppl_robustness}
\end{figure*}

\subsection{More Comparisons with Multimodal Baselines}
\label{appendix:more_comparisons}
We also compare our method with more multimodal,
i.e.,
audio-visual,
baselines under the cross-dataset setting.
Since they do not have open source codes or pre-trained weights,
we provide their results on the FakeAVCeleb according to~\cite{yang2023avoid}.

\input{table/more_results}

As shown in \cref{table:more_comparisons},
our method significantly outperform both supervised and unsupervised counterparts.
We note that many audio-visual face forgery detection methods adopt the cross-modal fusion strategy~\cite{zhou2021joint,yang2023avoid,oorloff2024avff},
However,
our method prove the cross-modal fusion may not be necessary for the face forgery detection task,
which we believe needs to be explored further.

\subsection{Robustness Experiments}
\label{appendix:robustness}
Following \cite{haliassos2021lips},
we apply the perturbations using the DeeperForensics \cite{jiang2020deeperforensics} code\footnote{\url{https://github.com/EndlessSora/DeeperForensics-1.0/tree/master/perturbation}},
which implements seven different perturbations conditioned on five intensity levels.
And we present some perturbation examples in \cref{fig:suppl_robustness}.

\input{table/suppl_robustness}

In \cref{table:suppl_robustness},
we report the average AUC scores across all intensity levels for each perturbation.
It can be seen our method achieves the state-of-the-art robustness to unseen perturbations,
although it starts from a lower AUC score than other supervised methods,
\eg, LipForensics \cite{haliassos2021lips} and FTCN \cite{zheng2021exploring}.

\begin{figure*}[h]
    \centering
    \includegraphics[width=0.78\linewidth]
    {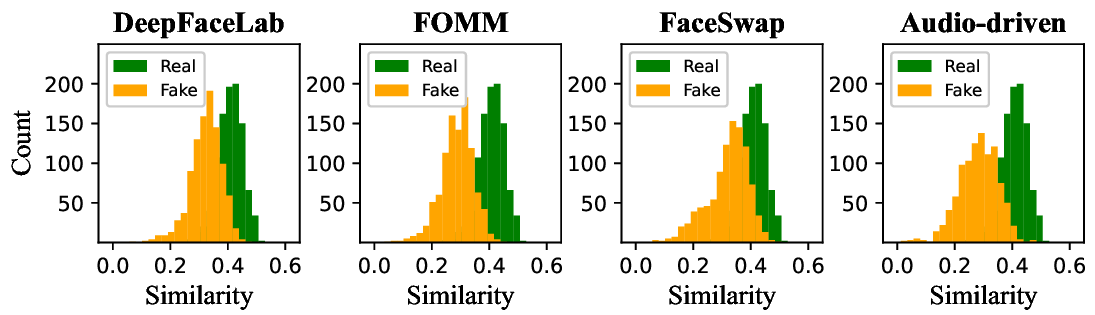}
    \caption{\textbf{Visualization of cosine similarity distribution of KoDF.}
    Cosine similarity distribution of audio and visual speech representations of different types of videos in KoDF.
    }
    \label{fig:suppl_visualization}
\end{figure*}
\subsection{Visualization Analysis}
\label{appendix:visualization}
In \cref{fig:suppl_visualization},
we show the cosine similarity distribution of different types of forgeries of KoDF \cite{kwon2021kodf},
involving DeepFaceLab \cite{perov2020deepfacelab}, FOMM \cite{siarohin2019first}, FaceSwap \cite{deepfakesfaceswapurl} and Audio-driven (including ATFHP \cite{yi2020audio} and Wav2Lip \cite{prajwal2020lip}).
It shows our method has strong generalization to unseen datasets and languages.

%% file: table/suppl_visual_frontend.tex
\begin{wraptable}{r}{6.5cm}
\vspace{-1.5em}
\caption{\textbf{Visual frontend architecture.}The output size is of the form $T\times H\times W$, where $T$ denotes the number of input frames, \textit{H} denotes the height of frames and $W$ denotes the width.}
\label{table:suppl_visual_frontend}
\begin{center}
\resizebox{\linewidth}{!}{
\begin{tabular}{c | c | c}
stage & filters & output size  \\ \toprule
conv\textsubscript{1} & $5\times 7\times 7$, stride $1\times 2\times 2$ & $T\times 44\times 44$ \\ \midrule
pool\textsubscript{1} & max, $1\times 3\times 3$, stride $1\times 2\times 2$ & $T\times 22\times 22$ \\ \midrule
res\textsubscript{1} & $\begin{bmatrix} 3\times 3, 64 \\ 3\times 3, 64 
\end{bmatrix} \times 2$ & $T\times 22\times 22$ \\ \midrule
res\textsubscript{2} & $\begin{bmatrix} 3\times 3, 128 \\ 3\times 3, 128 
\end{bmatrix} \times 2$ & $T\times 11\times 11$ \\ \midrule
res\textsubscript{3} & $\begin{bmatrix} 3\times 3, 256 \\ 3\times 3, 256 
\end{bmatrix} \times 2$ & $T\times 6\times 6$ \\ \midrule
res\textsubscript{4} & $\begin{bmatrix} 3\times 3, 512 \\ 3\times 3, 512 
\end{bmatrix} \times 2$ & $T\times 3\times 3$ \\ \midrule
pool\textsubscript{2} & global spatial average pool & $T\times 1\times 1$ \\ \midrule
\end{tabular}
}
\end{center}
\vspace{-2.5em}
\end{wraptable}

%% file: table/more_results.tex
\begin{wraptable}{r}{8cm}
\vspace{-0.5em}
\caption{\textbf{Cross-dataset generalization comparisons with multimodal baselines.}
We report the AUC (\%) scores on the FakeAVCeleb.}
\label{table:more_comparisons}
\centering
\resizebox{8cm}{!}{
\begin{tabular}{c c c c c c }
\toprule
Method & MDS~\cite{chugh2020not} & VFD~\cite{cheng2023voice} & Avoid-DF~\cite{yang2023avoid} & AVAD~\cite{feng2023self} & Ours \\  
\midrule
 & Supervised & Supervised & Supervised & Unsupervised & Unsupervised \\
 \midrule
AUC & 76.7 & 82.5 & 85.8 & 85.0 & \textbf{99.0} \\
\bottomrule
\end{tabular}
}
\vspace{-0.5em}
\end{wraptable}

%% file: table/suppl_robustness.tex
\begin{table*}[t]
\caption{\textbf{Average robustness to unseen perturbations.}
The average AUC (\%) scores of different methods over each type of perturbations at all intensity levels.
\textbf{Avg:}~average performance on corrupted videos. \textbf{Drop:}~decreased performance compared to clean videos.}
\label{table:suppl_robustness}
\begin{center}
\resizebox{0.98\linewidth}{!}{
\begin{tabular}{l c c c c c c c c c}\hline
Method & Clean & Saturation & Contrast & Block & Noise & Blur & Pixelation & Compress & \textbf{Avg/Drop} \\
\midrule
Xception \cite{rossler2019faceforensics++} & \textcolor{gray}{92.2} & 92.8 & 91.1 & 92.5 & 51.9 & 73.8 & 79.1 & 63.3 & 77.8/-14.4 \\
Face X-ray \cite{li2020face} & \textcolor{gray}{97.7} & 95.1 & 85.8 & \textbf{97.3} & 51.7 & 62.9 & 86.8 & 52.8 & 76.1/-21.6 \\
LipForensics \cite{haliassos2021lips} & \textcolor{gray}{99.6} & 99.2 & \textbf{99.6} & 83.4 & 77.0 & \textbf{98.7} & 94.0 & 63.5 & 87.9/-11.7 \\
FTCN \cite{zheng2021exploring} & \textcolor{gray}{99.4} & \textbf{99.3} & 95.2 & 78.2 & 52.0 & 89.6 & 81.0 & 85.7 & 83.0/-16.4\\
RealForensics\cite{haliassos2022leveraging} & \textcolor{gray}{99.6} & 98.7 & 98.7 & 76.8 & 75.6 & 98.3 & \textbf{98.5} & \textbf{97.2} & 92.0/-7.6 \\ 
Ours & \textcolor{gray}{97.6} & 97.4 & 96.7 & 94.6 & \textbf{80.8} & 97.4 & 96.7 & 94.2 & \textbf{94.1/-3.5}\\
\bottomrule
\end{tabular}
}
\end{center}
\end{table*}